\documentclass[conference]{IEEEtran}
\IEEEoverridecommandlockouts
\usepackage{cite}
\usepackage{amsmath,amssymb,amsthm, amsfonts}
\usepackage{algorithmic}
\usepackage{float}
\usepackage{graphicx}
\usepackage{textcomp}
\usepackage{xcolor}
\usepackage{tabularx}
\usepackage{soul}

\usepackage[utf8]{inputenc}
\usepackage[longend]{algorithm2e}
\usepackage{textgreek}
\usepackage{amssymb}
\usepackage{tcolorbox}
\usepackage{graphicx,array,caption}
\usepackage{enumitem}
\usepackage{lipsum} 

\makeatletter
\newcommand{\removelatexerror}{\let\@latex@error\@gobble}

\makeatother

\def\BibTeX{{\rm B\kern-.05em{\sc i\kern-.025em b}\kern-.08em
    T\kern-.1667em\lower.7ex\hbox{E}\kern-.125emX}}
\begin{document}

\title{Symphony: A Heuristic Normalized Calibrated Advantage Actor and Critic Algorithm with Application to Humanoid Robots
}

\author{
Timur Ishuov\textsuperscript{1}, 
Michele Folgheraiter\textsuperscript{2},
Madi Nurmanov,
Goncalo Gordo\textsuperscript{3},
Richárd Farkas\textsuperscript{1}, 
József Dombi\textsuperscript{1} \\
\textsuperscript{1}Department of Computer Algorithms and Artificial Intelligence, University of Szeged, Szeged, Hungary \\
\textsuperscript{2}Department of Robotics and Mechatronics, Nazarbayev University, Astana, Kazakhstan \\
\textsuperscript{3}Tinker AI Limited, London, UK \\

Emails: timur@inf.u-szeged.hu, michele.folgheraiter@nu.edu.kz, tinkerai.project@gmail.com\\

rfarkas@inf.u-szeged.hu, dombi@inf.u-szeged.hu
}

\maketitle

\begin{abstract}

In our work we implicitly suggest that it is a misconception to think that humans learn fast. The learning process takes time. Babies start learning to move in the restricted fluid environment of the womb. Children are often limited by an underdeveloped body. Even adults are not allowed to participate in complex competitions right away. However, with robots, when learning from scratch, we often don't have the privilege of waiting for tens of millions of steps. "Swaddling" regularization is responsible for restraining an agent in rapid but unstable development penalizing action strength in a specific way not affecting actions directly.

The Symphony, Transitional-policy\footnote{algorithm in between on and off policy due to balanced updates} Deterministic Actor and Critic algorithm, is a concise combination of different ideas for possibility of training humanoid robots from scratch with Sample Efficiency, Sample Proximity and Safety of Actions in mind. It is well known that a continuous increase in Gaussian noise without appropriate smoothing is harmful for motors and gearboxes. Moreover, an added noise trail, when actions are clipped, can accumulate at the action limits, potentially increasing the frequency of maximum control values (torque, velocity, angles). Compared to Stochastic algorithms, we use a different approach, we set a limited parametric noise and adjust the strength of actions instead of noise. Promoting a reduced strength of actions, we safely increase entropy, since the actions are submerged in weaker noise. When actions require more extreme values, actions rise above the weak noise and become more deterministic. Training becomes empirically much safer for both the surrounding environment and the robot's mechanisms. 

We use Fading Replay Buffer: using a fixed formula containing the hyperbolic tangent, we adjust the batch sampling probability: the memory contains a recent memory and a long-term memory trail. Fading Replay Buffer with increased Update-To-Data ratio (3) allows us to use Temporal Advantage when we improve the current Critic Network prediction compared to the exponential moving average. Temporal Advantage allows us to update the Actor and the Critic in one pass, as well as combine the Actor and the Critic in one Object (Class) and implement their Losses in one line. The Symphony algorithm is called Heuristic\footnote{a hypothetical approach of combining different methods and using specific parameters based on intuition and experience to solve practical problems} and Calibrated, since the activation, loss, regularization functions were rewritten and parameters were customized during numerous experiments in order to bring safe model-free imitation-less Humanoid robot training closer to reality.

\end{abstract}


\section{Preliminary}

We consider an Agent or Actor that utilizes a Policy $A_{\phi}$\ or Actor Network with parameters  $\phi$\ which at a Continuous Markov Decision Process (MDP) State (hereinafter referred to as simply "state") $s_t$ produces a continuous vector of actions (hereinafter "actions") $a_t+\epsilon \in [-a_{max}, a_{max}]$. The Environment responds with reward $r_t$ and the Agent appears at the Next MDP State (next state) $s_{t+1}$. We use the terms Critic, Critic Network, and Q Network interchangeably, while the Critic's parameters are abbreviated by $\theta$, and $det$ or * abbreviation is used for values detached from the computation graph to compute gradients.

\section{Introduction}

The Symphony algorithm aims to solve two conflicting problems of Model-Free Imitation-Less Reinforcement Learning: Sample Efficiency and harmonious agent motion without jerky movements, which is achieved with the help of Sample Proximity and Safety of Actions. On-policy algorithms like Proximal Policy Optimization\cite{schulman2017proximal} (PPO) solved the problem of harmonious agent movements using a small “safe” gradient step, which leads to a larger number of close samples and better policy consequently. However, PPO has proven effective in simulations, where the learning process can be parallelized across many virtual environments using Graphics Processing Units or Tensor Processing Units, and make a gradient step based on a larger number of roll-outs, so this process requires a careful transition from simulation to a real robot, though no bootstrapping (updating Online Q Network based on prediction of its delayed variant) is used. When it comes to training on a real robot from scratch, entropy regularization based off-policy sample-efficient algorithms like Soft Actor- Critic\cite{haarnoja2018soft} (SAC) and its advanced derivatives are often used, though this family of algorithms can suffer from prediction inaccuracies of the TD equation; the regularization mitigates this issue to some degree.

To achieve even higher Sample-Efficiency of the SAC algorithm, recent studies (SAC-20, Randomized Ensembled Double Q learning or RedQ \cite{chen2021randomized}) have concluded that it is possible to do a larger number of updates-per-frame. Though taking the minimum of two prediction functions of the future aggregated reward min(Q1, Q2)\cite{fujimoto2018addressing} at an increased update frequency increases the probability of stagnation of the policy or Actor Critic at a local extremum due to the constantly decreasing step of min(Q1, Q2); Random Q network selection from Q network ensemble (RedQ algorithm) helps against this. The Aggressive Q learning with Ensembles\cite{wu2021aggressive} (AQE) algorithm instead of min(Q1, Q2) uses the idea of Truncated Quantile Critics\cite{kuznetsov2020controlling} (TQC) which is about taking the average in a larger number of Q heads or nodes (Q distribution) after truncating several highest Q nodes. Truncation is necessary due to the overestimation and high variability of the average value (for Walker-2d this is up to half of the nodes). This step allows achieving highest scores. Another work (Reset\cite{nikishin2022primacy} algorithm) suggests leaving the minimization but refreshing the weights of the last two layers to the default ones, periodic refresh allows using a higher update frequency, up to 128.

If we look at the work of Distributional Soft Actor and Critic\cite{ma2020dsac}, we can understand that the average value is a kind of "Pandora's box" for humanoid agent movements by increasing the variation and the number of attempts or roll-outs\cite{bellemare2017distributional}. The point is that when we take the minimum of two functions for predicting the best actions, with time we create a very careful, cautious or risk-averse agent, but when we take the average, we give the agent the opportunity to be relatively optimistic in its predictions and make more mistakes; for humanoid agents, this means more falls. But when the agent falls more, this in turn leads to better experience, which results in more harmonious movements and more deterministic  $Q=r_{done}$ in the batch sample, which in the Temporal Difference equation provides better predictions in the end. Unfortunately, this takes some time, while introducing a high instability in the learning process - the first version of Distributional Soft Actor and Critic still required several millions of steps. Most likely, because of this, the authors of this algorithm returned to the minimum between two Q distributions\cite{duan2023dsac}

\section{Foundation}

The Symphony is an advantage based transitional-policy deterministic Actor-Critic algorithm with limited\footnote{The parametric standard deviation of $1/e$ scales Standard Gaussian noise's range to [-1, 1]} Normal Gaussian noise ($\epsilon \sim a_{max}*1/e*\mathcal{N}(0.0,\,1.0)$, where $\mathcal{N} \in [-e, e]$) added to output actions $a$.

Symphony increases Temporal Advantage (TA) or plain difference between a currently predicted target Q value and exponential weighted moving average\footnote{the exponential weighted moving average of Q target value has some resemblance to the Approximate Value Function, though the latter uses the return traces R for the average $Q_\pi(s,a)= (1-\gamma)  R_t(s,a) + \gamma \, V(s_t|\psi)$}\cite{Roberts01081959} of target Q values (copy of Q values, $Q^*$, detached from the gradient computation) with heuristic $\alpha=0.5..0.7$.

The Symphony's Objective Function can be expressed as:

\begin{equation}
\mathbb{E}_{s_t\sim\rho^\beta}[\nabla_{a} \left[Q_{target}(s_{t+1},a(s_{t+1}))-Q_{T}\right]\nabla_{\phi}a(s_{t+1}|\phi)]
\end{equation}

where $Q_{T} = \alpha \: \bar{Q}_{T-1} + (1-\alpha)  \: Q^*_{target}$.

\;

In Symphony, instead of ensemble of 5-10 networks or only 1-2 Critics, we take 3 Critics each of which outputs 128 nodes and  concatenate output nodes together producing a distribution of size 384. For the output of the Target Critic, we sort the distribution from minimum to maximum and multiply those values by fixed weights that are opposite to the Fading Replay Buffer probabilities (which will be discussed further) though for the length of 384. The weighted sum of 384 Critic's nodes becomes the output of the Target Critic.

\begin{table}
\caption{Fixed weights for Target Critic's nodes}
\begin{tabular}{m{0.18\textwidth}m{0.28\textwidth}}

\includegraphics[width=0.9\linewidth]{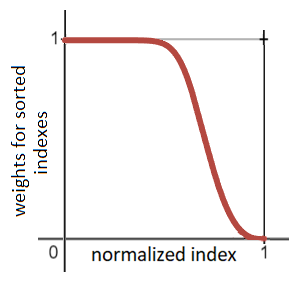} &
\begin{enumerate}[itemsep=5pt]
 \item    $i_n = \frac{\{1,2,3...384\}}{384}$ 
 \item    $w_{i_n} = tanh((\pi(1-i_n))^{e})$
  \item    $w_n =  w_{i_n}/\sum_{1}^{384}w_{i_n}$ 
\end{enumerate} \\
\\
\multicolumn{2}{c} {$Q_{target} = \sum_{n=1}^{384} (w_n * q_n)$}\\

\end{tabular}
\end{table}

Though, Target Critic based Actor's updates can slow down training due to Polyak Averaging\cite{Ruppert} \cite{Polyak} via soft updates ($\theta^{T} \leftarrow \tau\theta + (1-\tau)\theta^{T}$)\cite{lillicrap2015continuous}, Advantage or $\Delta Q$ gives us some independence from the actual Q values, and together with the increased update-to-data ratio $\mathcal{G}$ (Symphony uses 3 updates per frame) this method provides a "seamless" computationally efficient update of the Actor-Critic. For this we predict the next action $a^{'}_{t+1}$ using the next state and Online Actor ($a^{'}_{t+1} = A_{\phi}(s_{t+1})$), then use the next state and the next action to produce next target $Q$ value and its detached variant:

\begin{equation}
Q_{target}, Q_{target}^{*} = Q_{target}(s_{t+1},a^{'}_{t+1})
\end{equation}

which are used to update simultaneously Online Actor and Online Critic using Temporal Advantage and Temporal Difference\cite{Sutton1988}, respectively:

\begin{equation}
- L_{\phi}(\frac{Q_{target}-Q_{T}}{|Q_{T}|}) + L_{\theta}(\widehat{r_t} +\gamma Q_{target}^{*} - Q_{online})
\end{equation}

 Simultaneous update of Online Networks becomes possible when we combine the Actor and Critic into a single Class, or unified Object consequently. We do not utilize Target Actor, while the Target Critic is updated using Polyak Averaging as previously mentioned.

\subsection*{Equation revision ...}

One can notice $|Q_{T}|$ in the denominator of TA, which is used for Advantage normalization, while  $\widehat{r_t}$  are normalized rewards. Normalization is performed using $\bar{r}_{n}$, a mean of absolute values of the exploratory rewards $r_{exp}$ collected during $N^{exp}=10,240$ exploration steps:
\begin{equation}
\bar{r}_{n} = \frac{1}{N_{exp}}\sum_{1}^{N_{exp}}|r_{exp}|
\end{equation}

 Normalization is done after exploration and further using estimated $\bar{r}_{n}$.

\subsection*{... end of revision}

As the Replay Buffer grows in size, sampling evenly with a small batch size (e.g., 32–64) may lead to Q-values predicted from one batch having a very weak correlation with Q predictions from the other. To address this, we use the previously mentioned exponential smoothing, an increased batch size of 384, and a Fading Replay Buffer, where sampling priorities $p$ correspond to the normalized indexes of the transitions $i$ (normalized by the capacity of the Replay Buffer, $N_{rb}$).

More specifically, we smoothed linear priorities with $tanh((\pi i)^{e})$ giving a smooth transition (by applying $\pi$ to the input of the hyperbolic tangent function, we compress the function’s transition region closer to the 0..1 range) and plateau like shape for the latest experience (order $e$). To obtain probabilities that add up to 1.0 we further divide weights/priorities by a sum of all weights/priorities.

\begin{table}[ht]
\centering
\caption{Probabilities for the Fading Replay Buffer}
\begin{tabular}{c}

\\
\normalsize
1) $ i_n = \frac{\{1,2,3...N_{rb}\}}{N_{rb}}$ \\[1.5em]
\normalsize
2) $w_{i_n} = tanh((\pi i_n)^{e})$ \\[1.5em]
\normalsize
3) $p = \frac{w_{i_n}}{\sum_{1}^{N_{rb}}w_{i_n}}$ \\[1.5em]
\\
\includegraphics[width=0.9\linewidth]{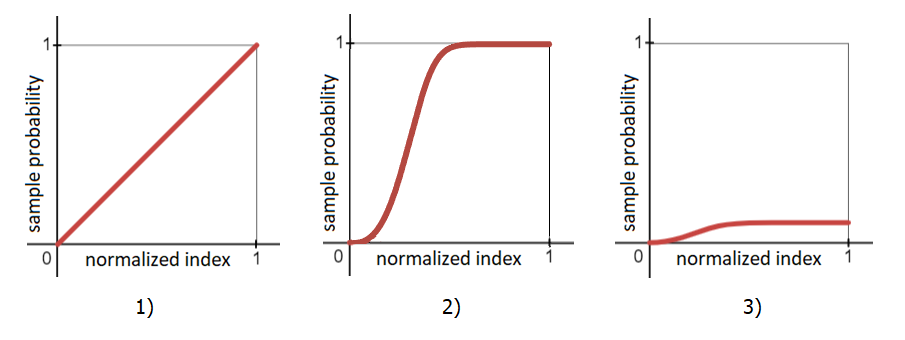} \\
\end{tabular}
\end{table}

Finally, we repeat 10,240 Exploration transitions 50 times, creating a fully filled Replay Buffer of length 512,000 from the beginning. In this way we ensure that all Exploration transitions will be thoroughly processed before disappearing. After the Replay Buffer is filled, a new transition enters it at the last index. Then we do roll or shift left operation so that the first (oldest) transition appears at the last index and is ready to be overwritten. When we encounter a transition that contains $\widehat{r_t}=r_{done}$ at the oldest placeholder, we do shift left 2 positions instead of 1, so that Terminal points are never overwritten. Furthermore, outside of the Replay Buffer, when agent learns to move without falling or approaches an artificial termination limit, e.g. 1000, at a timestep close to termination (950) we zero the actions $a_t$ and leave only noise, so that the agent falls deliberately. This resembles human memory which remembers pivotal points and makes the Actor-Critic Network less dependent on Bootstrapping.

The Fading Replay Buffer is an intermediate step between On-policy Cache type Memory and Off-policy Experience Replay\cite{Lin1992SelfimprovingRA}. It smooths difference between both memories since actions that were done far in the past have a little relevance to the current policy, but it is also important to improve Q from those old "bad" transitions for better generalization. Also Fading Replay Buffer diminishes a  transition problem of Experience Replay as a vanilla Experience Replay is a Stack type memory until it reaches its capacity then it is a Double-ended queue or a Standard Moving Average.

Temporal Advantage and Fading Replay Buffer introduce high sample efficiency which can lead to the fast growth in Q learning. However, if more closely related experiments lead to more harmonious agent movements but requires large number of samples, then how to achieve Sample Proximity and Sample Efficiency if these two goals contradict each other? 

It may seem counterintuitive but encouraging the agent not to use the full range of motion, "Swaddling" or Control Cost punishment, can help.
 
Before we learned to run, we learned to walk, before we learned to walk, we learned to stand. If a small child is not swaddled and not kept in a crib, he can cause a lot of trouble to himself and to the Environment. How is swaddling different from "caution" in predictions? In that variability is preserved, but in a small range. This means that before learning to run like Usain Bolt, our agent must learn to move its legs correctly. 

If we look at the reward formation in the Mujoco Humanoid-v4 environment, we subtract the sum of squares of actions from the positive direction of speed. That is, we punish the agent for the power or strength of actions – the stronger the actions, the stronger the punishment, although the weight of this punishment (control cost) is 0.1, and the weight of the speed reward is 1.25, and the healthy state is 5. It turns out that the relative "Swaddling" is embedded in the reward. But the weight value for the control cost in Walker-2d is only 0.001. When we do a comparative analysis using these environments, the environments themselves act as a black box, and we cannot influence the internal parameters of the environment (which does not apply to developing our own robot)
What is "Swaddling" regarding the algorithm? In the simplest implementation, it is a direct soft suppression of bigger actions by adding control cost to the Actor’s Loss Function:

\begin{equation}
ctrl\_cost = \beta \:a^2
\end{equation}

where $\beta$ is a temperature parameter, a - actions. $a^2$ in Loss function will reduce the strength of the actions depending on the $\beta$ coefficient. The $\beta$'s strength should be relatively small, otherwise we are unlikely to learn deterministic moves. Adjusting $\beta$, we facilitate actions that can be viewed as buried in limited Gaussian Noise increasing Entropy Level. The difference from a learnable standard deviation is that Noise Level is set constant, it will never go to unprecedented values.

With the introduction of a new harmonic activation function that naturally induces fluctuations in the output and loss functions that allow for a less strict penalization of stronger values, and using the principles behind these functions, instead of adding the control cost, we introduced a new approach called "Swaddling" which does not affect actions directly.

Harmonic activation functions are rarely used in Reinforcement Learning due to high levels of instability. But if stabilized they can introduce infinite exploration in a more natural way than hard resetting neural networks introduced in Reset mechanism\cite{nikishin2022primacy} as small input signal can cause some oscillatory effect (as in the muscular system of animals).

We use internal Sine wave function in between two Linear Layers both in Actor and Critic, which represents an approximation to a Fourier Series in the form of: $Asin(wx+b)+c$, where an absence of the Cosine part is to some extent covered by the presence of a learnable phase shift. A harmonic and non-linear nature of a Sine wave introduces instability and consequently a broader search for solutions in the learning process. We rectified those instabilities by a Swish or SiLU activation function with a weight factor of 1.5.

During experiments, we slightly decreased the scale of the Rectified Sine function, which showed better balance between exploration and exploitation. However, the best choice was to implement proportional scaling ($\sigma$) using a trainable vector s with a length equal to the hidden dimension (initialized uniformly between -$\frac{1}{\sqrt{dim}}$ and $\frac{1}{\sqrt{dim}}$). After this vector is squashed between 0 and 1 using a sigmoid function, it is then used to proportionally scale the activation function.

\begin{table}[ht]
\centering
\caption{Rectified Sine or ReSine}
\begin{tabular}{c}

\\
\normalsize
1) $ \sigma = \text{sigmoid}(s)$ \\[1.5em]
\normalsize
2) $f(x) = \sigma \sin\left(\frac{x}{\sigma}\right)$ \\[1.5em]
\normalsize
3) $F(x) = f(x) \text{ sigmoid}\left(\frac{1.5}{\sigma} f(x)\right)$ \\[1.5em]
\\
\includegraphics[width=0.9\linewidth]{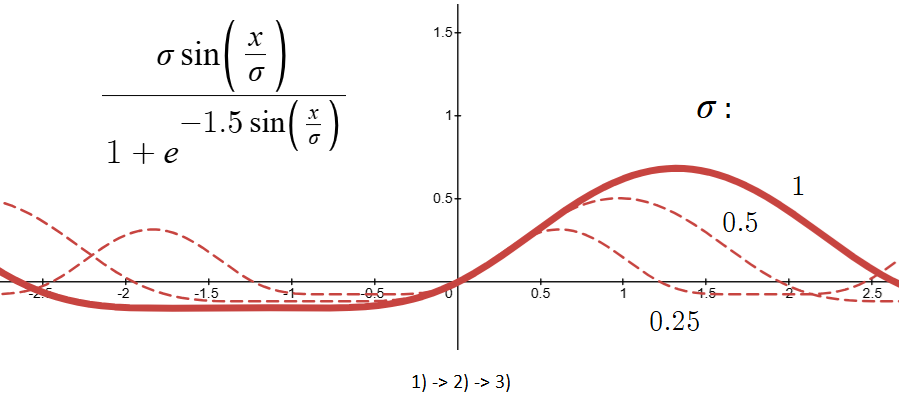} \\
\end{tabular}
\end{table}

Apart from stabilization, rectification additionally provides pseudo-random harmonic dropouts at rectified areas, while noise produced in the form of harmonics is visually closer to the movements of living beings.

We discussed Rectified Huber Symmetric $x\,tanh(x)$ and Asymmetric $|x|\,tanh(x)$ Error (ReHSE, ReHAE) functions previously\cite{9945743}\cite{Ishuov_Timur2023-10-11}. x is the error between  prediction and target for the Loss function (ReHSE) or prediction and baseline for the Advantage function (ReHAE).

They have almost quadratic relationships $x^2$ for relatively small values, but the functions strive for linearity for big values as tanh approaches 1 for bigger values. The difference with the Huber loss function is that transition happens gradually. Symmetric version ($x\,tanh(x)$) is used to increase stability in Critic learning, while faster learning can be achieved when the Asymmetric version ($|x|\,tanh(x)$) is used in conjunction with Advantage for Actor’s updates. For the Advantage it is important to keep the sign intact.

Delta or Advantage utilizes gradual change from quadratic to linear relationship most of the time, whereas Q value prediction even if it starts from low values initially, rapidly goes to stronger values ($\gg\,$1.0) for positive rewards. The benefit of quadratic relationship at smaller values is a dampening effect of small deterministic gradients. Additionally, we made ReHSE and ReHAE closer to the Huber representation by dividing the argument under tanh by 2: $xtanh(x/2)$ and $|x|tanh(x/2)$.

\begin{figure}[H]
    \centering
    \includegraphics[width=8cm]{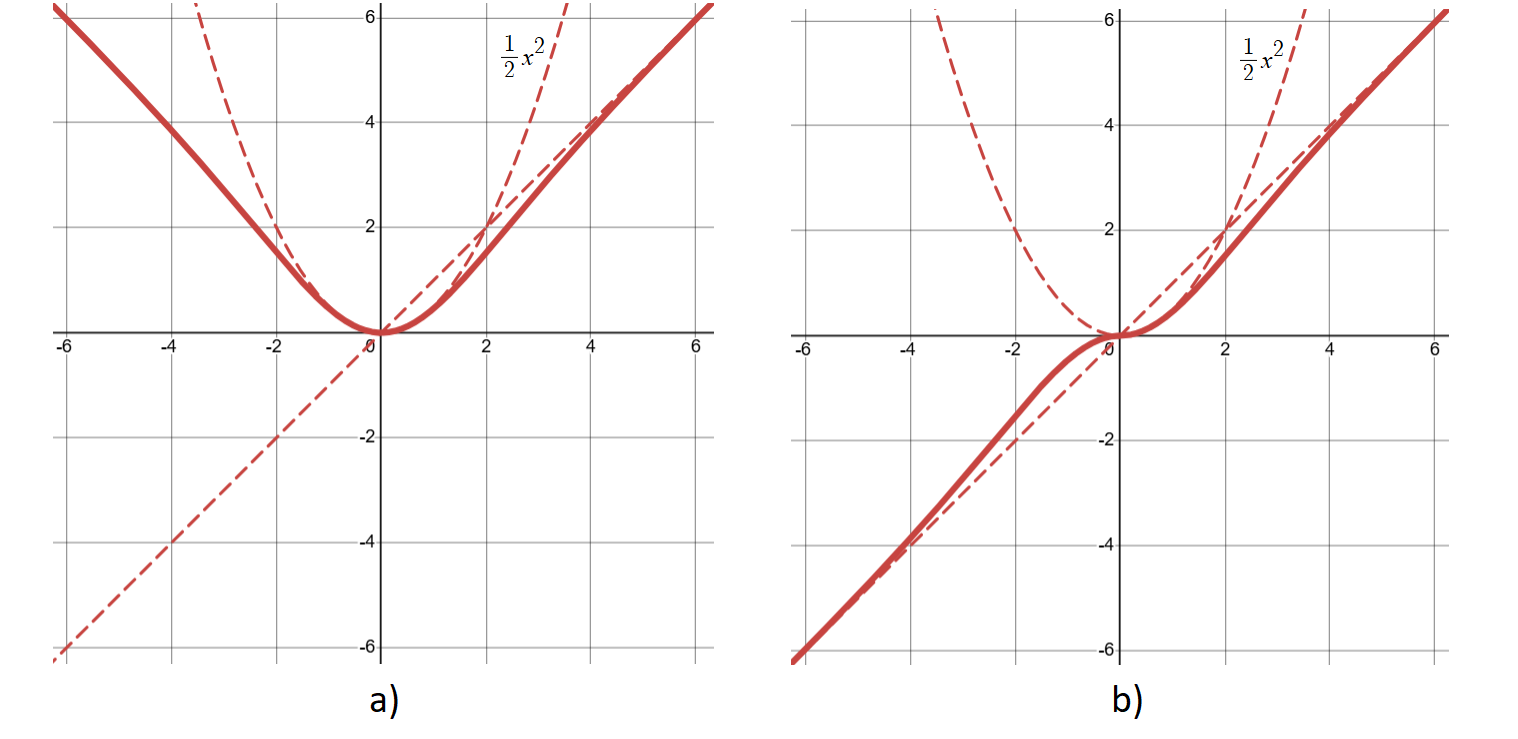}
    \caption[Caption for LOF]{a) x*tanh(x/2) - Rectified Huber Symmetric Error, b) abs(x)*tanh(x/2) - Rectified Huber Asymmetric Error.}
    \label{fig:ReHSE_ReHAE}
\end{figure}

\begin{figure}[H]
    \centering
    \includegraphics[width=8cm]{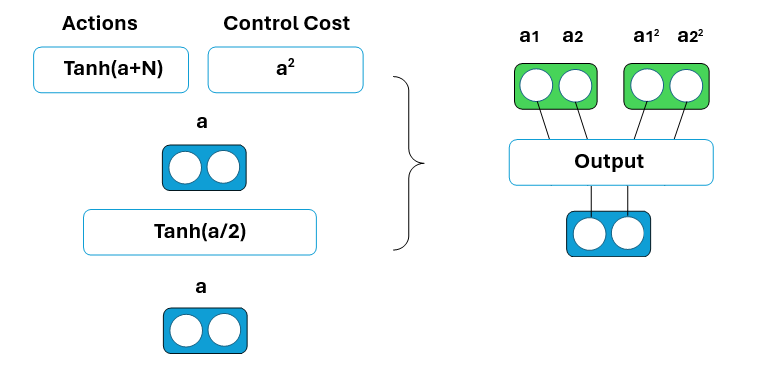}
    \caption[Caption for LOF]{"Control Cost" regularization.}
    \label{fig:control_cost}
\end{figure}

The expression $tanh(x/2)$ can also be used for a sparser squashing of actions before adding noise. And when noise is introduced, a more linear squashing using $tanh(x)$ instead of action clipping can help prevent noise accumulation at the action limits. At the action edges, Gaussian noise will become slightly skewed toward the center with a narrower distribution due to the applied tanh function. Considering action regularization, both processes can be combined into the representation shown in Figure~\ref{fig:control_cost}, where $N=1/e*\mathcal{N}(0.0,\,1.0)$, $\mathcal{N} \in [-e, e]$.

Unfortunately, the direct relationship between “Control Cost” regularization and actions can impose stronger constraint to the learning process, thus the temperature $\beta$ should be infinitely small.

\subsection*{Decoupling action regularization from actions}

\begin{figure}[H]
    \centering
    \includegraphics[width=8cm]{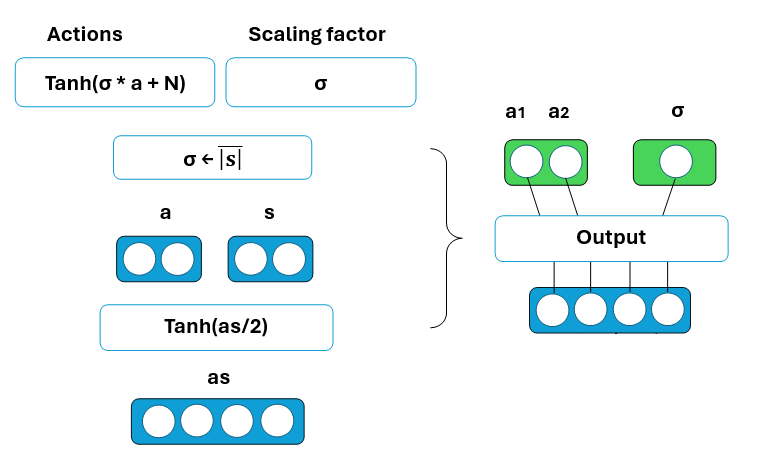}
    \caption[Caption for LOF]{Action decoupling process}
    \label{fig:Output2}
\end{figure}

We can use a relatively similar approach that we used in proportional scaling for the ReSine activation function. We can limit the action scaling using the limiting vector s. In this case, the Actor Network will output two independent parameters a and s, both with the action dimension. We can process s through the same $tanh(x/2)$ function. The only thing we need to do is to find mean over absolute values of vector s (the average counteracts overfitting), Figure ~\ref{fig:Output2}.

 For regularization of $\sigma$, we can use inverse tanh (ATanh) with argument $\sigma^{1/\beta}$ creating a barrier regularization for values approaching 1.0. Under regularization ATanh strives for its minimum value of 0.0, while $\beta$ responds for creation of a symmetric plateau around 0.0. To differentiate it from the control cost ($a^2$) we referred to it as "Swaddling" ($\Omega$)
 
We must multiply output actions by $a_{max}$, for simplicity we omitted this step in the figure. A factor 2 is used for Swaddling, as $2\,ATanh(x)$ is inverse for $Tanh(x/2)$.

\begin{figure}[H]
    \centering
    \includegraphics[width=8cm]{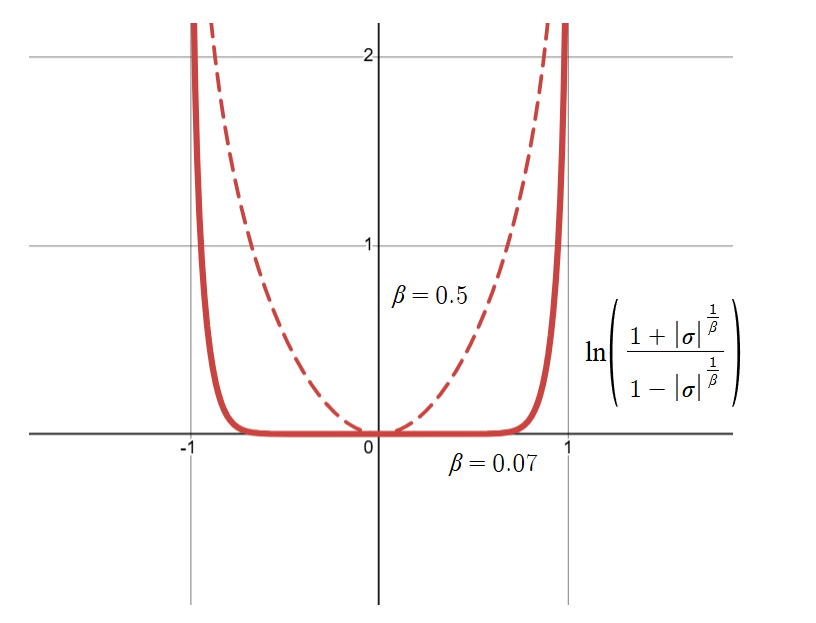}
    \caption[Caption for LOF]{Swaddling Regularization with $\beta$ as  temperature.}
    \label{fig:swaddling}
\end{figure}

\subsection*{Adjusting $\beta$}

When the degree $\beta=0.5$, the function $\Omega(x) = \ln{(\frac{1+x}{1-x})}$ (where $x\leftarrow|\sigma|^{1/\beta}$) resembles a hyperbola near 0.0, while it approaches +inf as $\sigma$ approaches 1.0. When we decrease $\beta$ we give a broader range for our scaling factor to operate with, e.g. $\beta=0.07$. 

To increase entropy, the $\beta$ value should suppress scaling factor values above the standard deviation of noise, but prevent the scaling factor from rolling toward 0.0.
To determine $\beta$ more precisely, we can use a nontrivial approach by adding a helper function $\omega(x)=x\,ln(x)$. This function's minimum is reached at $x=\frac{1}{e}$, precisely the value of the parametric standard deviation. However, if we add this function with the $\beta$ parameter to align it with ATanh, then for a strong beta, the minimum will shift closer to the center, while we want the scaling factor to float around the standard deviation. The combined equation is difficult to solve mathematically for $\beta$, but heuristically, we see that values $\beta\leq0.05$ provide the correct match (for values smaller than 0.05, the scale value for the function's extremum coincides with 1/e):

\begin{figure}[H]
    \centering
    \includegraphics[width=9cm]{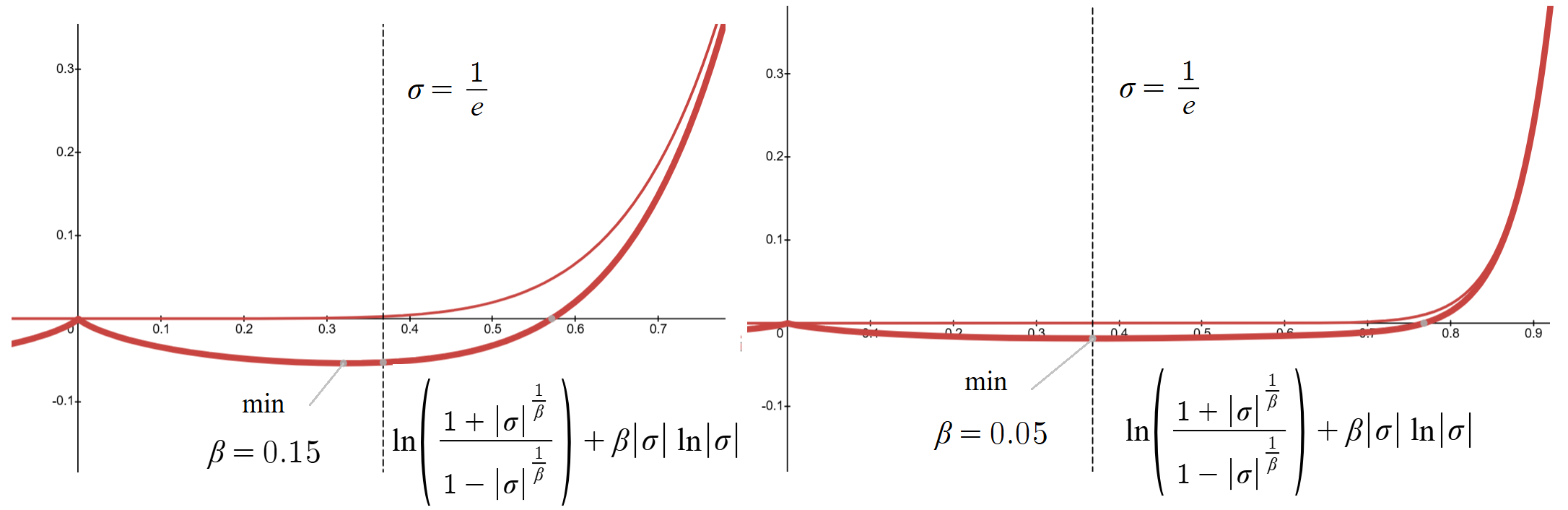}
    \caption[Caption for LOF]{Approximating $\beta$ value}
    \label{fig:Swaddling2}
\end{figure}

For adjustable Beta, it is possible to make it trainable. We repeated the previous approach where the Actor Network outputs several parameters, now $\sigma$ and $\beta$ additionally to actions, except we do not find means of absolute values of vectors s and b because $\sigma$ and $\beta$ are to balance each other to some extent, though they are clipped between $1\mathrm{e}{-3}$  and $1.0 - 1\mathrm{e}{-3}$:

\begin{figure}[H]
    \centering
    \includegraphics[width=8cm]{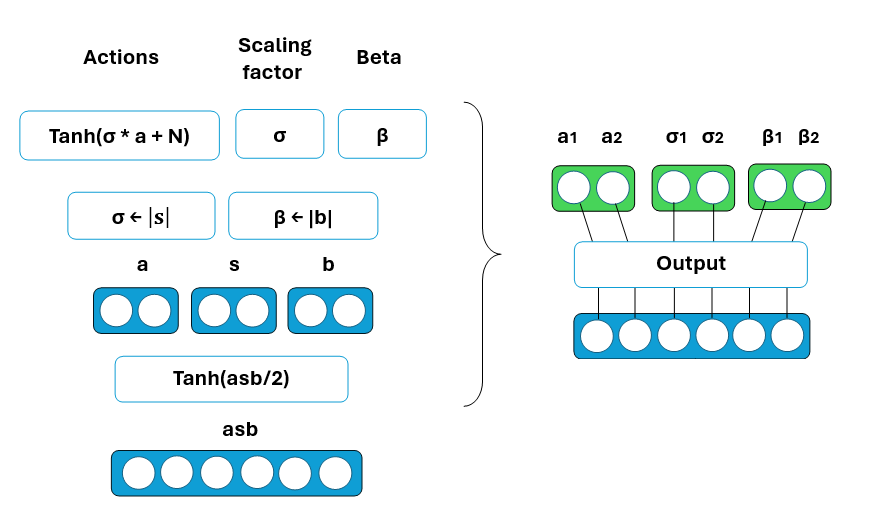}
    \caption[Caption for LOF]{$\beta$ as trainable parameter}
    \label{fig:Output5}
\end{figure}

$2\,ATanh(x)$ is also used to regularize Beta, with the function argument being $\beta^{1/0.5}$ or $\beta^2$.  A separate helper function for Beta is not needed, since $\beta$, by the defined property, should converge closer to 0.0. While scaling factor $\sigma$ participates in Q value learning through produced actions, $\beta$ mostly is a resisting factor.  Sending $\beta$ without power of 2 will result in a higher gradient value directed towards 0.0 which quickly diminishes $\beta$; while sending $\beta$ with a higher order in opposite decreases the gradient and imposes a stronger obstacle for Q value learning through increased $\beta$. On the one hand the helper function already steers Beta toward 1.0 (function's minimum), on the other the additional hyperbola shaped regularization suppresses its value (we detach $\beta$ from the gradient computation for the original Swaddling Function):

\begin{figure}[H]
    \centering
    \includegraphics[width=7cm]{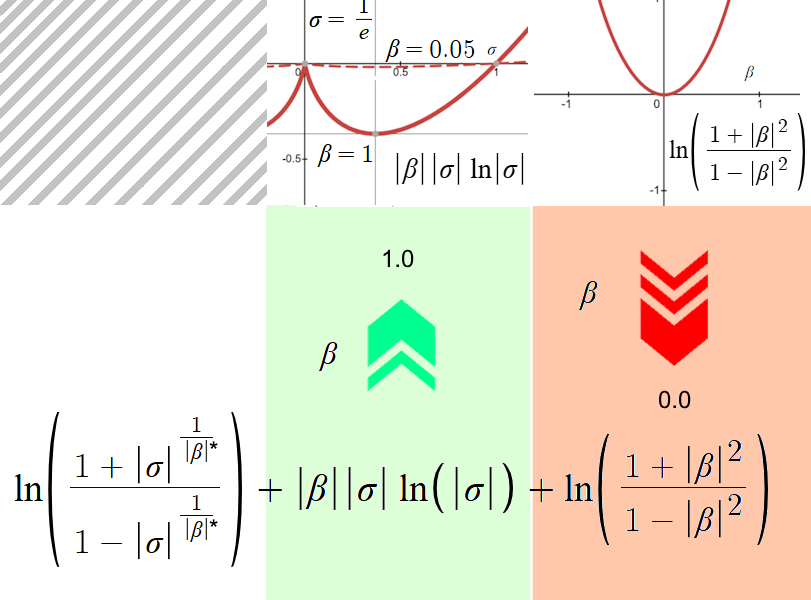}
    \caption[Caption for LOF]{Full Swaddling Regularization}
    \label{fig:Swaddling3}
\end{figure}

\subsection*{Hyper-parameters matter}

In Symphony algorithm we rigidly preset some constants that can also effect the final performance. 

The reciprocal of the golden ratio for the smoothing factor of the TA's exponential moving average $\alpha=\frac{1}{\phi}=\frac{\sqrt{5}-1}{2}\approx0.62$, reward normalization factor $\bar{r}_n= \overline{|r_{exp}|}$ calculated after $N_{exp}=10,240$ exploratory steps, which are repeated 50 times to produce a memory of 512,000 length from the beginning. 

The Fading Replay Buffer's capacity of $N_{rb}=50\,N_{exp}$, decay factor $\gamma=0.99$, delay constant $\tau=0.005$, Update-To-Data $\mathcal{G}=3$.

The Actor-Critic network's Optimizer is AdamW\cite{kingma2017adam}\cite{loshchilov2019decoupledweightdecayregularization} with the learning rate $\alpha_{lr}=1\mathrm{e}{-4}$, less sharp gradient with $\beta1,\beta2=(\alpha, 1-\tau)\approx(0.62, 0.995)$ and weight decay $\lambda=0.01$. Performance optimization: we removed Adam's\cite{kingma2017adam} bias correction as it is not essential in the Reinforcement Learning setup; the weight decay complements the main gradient value without a significant alteration of AdamW's principles: 

\begin{equation}
\theta_{t}\leftarrow\theta_{t-1}-\alpha_{lr}\,(m_t/(v_t+\epsilon) + \lambda\,\theta_{t-1})
\end{equation}

We use Layer Normalization\cite{ba2016layer} after the first fully connected layer. The hidden layer size $h_{dim}$ is $N_{rb}\div1000=512$ for Actor-Critic inner networks. Each Critic Network has $h_{dim}\div 4 = 128$ output nodes. The batch-size $\mathcal{B}=384$, the same as the number of concatenated output nodes of the Critic Network.

\begin{tcolorbox}[colback=white!5!white,colframe=black!75!black,fonttitle=\bfseries, title=Symphony: Pseudo-code]
\removelatexerror
\begin{algorithm}[H]
Initialize Actor-Critic Network ($A_{\phi}$, $Q_{\theta}$)\;
Initialize Target Critic Network $Q_{\theta^{T}}$ ($\theta^{T} \leftarrow \theta$)\;
Initialize Fading Replay Buffer $\mathcal F$($N_{rb}$)\;
Set default $\alpha_{lr}, \gamma, \tau,  \mathcal{B}, \mathcal{G}$. Set  $\alpha=\frac{1}{\phi}$\;

After $N_{exp}$ estimate $\bar{r}_n$ and normalize $r_{exp}$\;
Fill $\mathcal F$ by repeating $N_{exp}$\ transitions 50 times, 
update priorities in $\mathcal F$\;

\ForEach{episode}{
    \ForEach{step of episode}{
        \:\\
        Take an action $a_t$, observe $r_t$, $s_{t+1}$\;
        Store $i$ transition ($s_t,a_t,r_t,s_{t+1}$) in $\mathcal F$\;
        Sample $\mathcal{B}$ ($s_t,a_t,r_t,s_{t+1}$)  transitions  using priorities from $\mathcal F$\;
        \ForEach{$\mathcal{G}$ updates}{
        \:\\
        
        $a^{'}_{t+1}, \sigma_{t+1}, \beta_{t+1} = A_{\phi}(s_{t+1})$\\
        \:\\

        $Q_{\theta^{T}}, Q_{\theta^{T}}^{*} = Q_{\theta^{T}}(s_{t+1},a^{'}_{t+1})$\\

        \:\\

        $Q_{T} = \alpha\:\bar{Q}_{T-1} + (1-\alpha)\: Q_{\theta^{T}}^{*}$\\
        \:\\

        $L_{\phi,\theta} =  -\mathrm{ReHAE}(\frac{Q_{\theta^{T}} - Q_{T}}{|Q_{T}|})$\\
                  \:\\

        $  +\,\mathrm{ReHSE} ( \widehat{r_t}+\gamma Q_{\theta^{T}}^{*} - Q(s,a))  $\\

    \:\\

         $ +\,\Omega\omega\,(\sigma_{t+1}, \beta_{t+1})$\\
        \:\\

        $\phi \leftarrow \phi - \alpha_{lr} \frac{\partial  \bar{L}_{\phi,\theta}}{\partial \, \phi} $\\ 
        \:\\
        $\theta \leftarrow \theta -\alpha_{lr} \: \frac{\partial \bar{L}_{\phi,\theta}} {\partial \, \theta} $\\
        \:\\

        $\theta^{T} \leftarrow  (1-\tau)\theta^{T} + \tau\theta $\\
        \:\\
        
        }
        Change random seeds\;
        \:\\
    }
}
\_\_\_\_\_\_\_\_\_\\
\footnotesize{

\:\\

\renewcommand{\arraystretch}{1.2}

\begin{tabular}{ m{4em}|m{7em} | m{4em} |m{7em} }

$':$& predicted actions & $*:$& detached values \\

$\bar{r}_n:$ & $\overline{|r_{exp}|}$ & $\widehat{r_t}:$ & $r_t/\bar{r}_n$ \\

 ReHSE(x): & $x\,tanh(x/2)$ & ReHAE(x): & $|x|\,tanh(x/2)$  \\

  $\Omega(x):$ & $ln(\frac{1+x}{1-x})$ & $\omega(x):$ & $x\,ln(x)$ \\
 
$\Omega\omega(x,\kappa):$ & \multicolumn{3} {l}{$ \Omega\,(x^{1/\kappa^{*}}) + \kappa\,\omega(x) + \Omega\,(\kappa^2)$}  \\

\end{tabular}
\:\\
Under ReHSE, ReHAE, $\Omega\omega$ functions mean is calculated over batch of size $\mathcal{B}$. Standard Gradient Descent used as a shorthand for customized AdamW optimizer\;
}

\end{algorithm}
\end{tcolorbox}

\begin{figure*}[h]
    \centering
    \includegraphics[width=17cm]{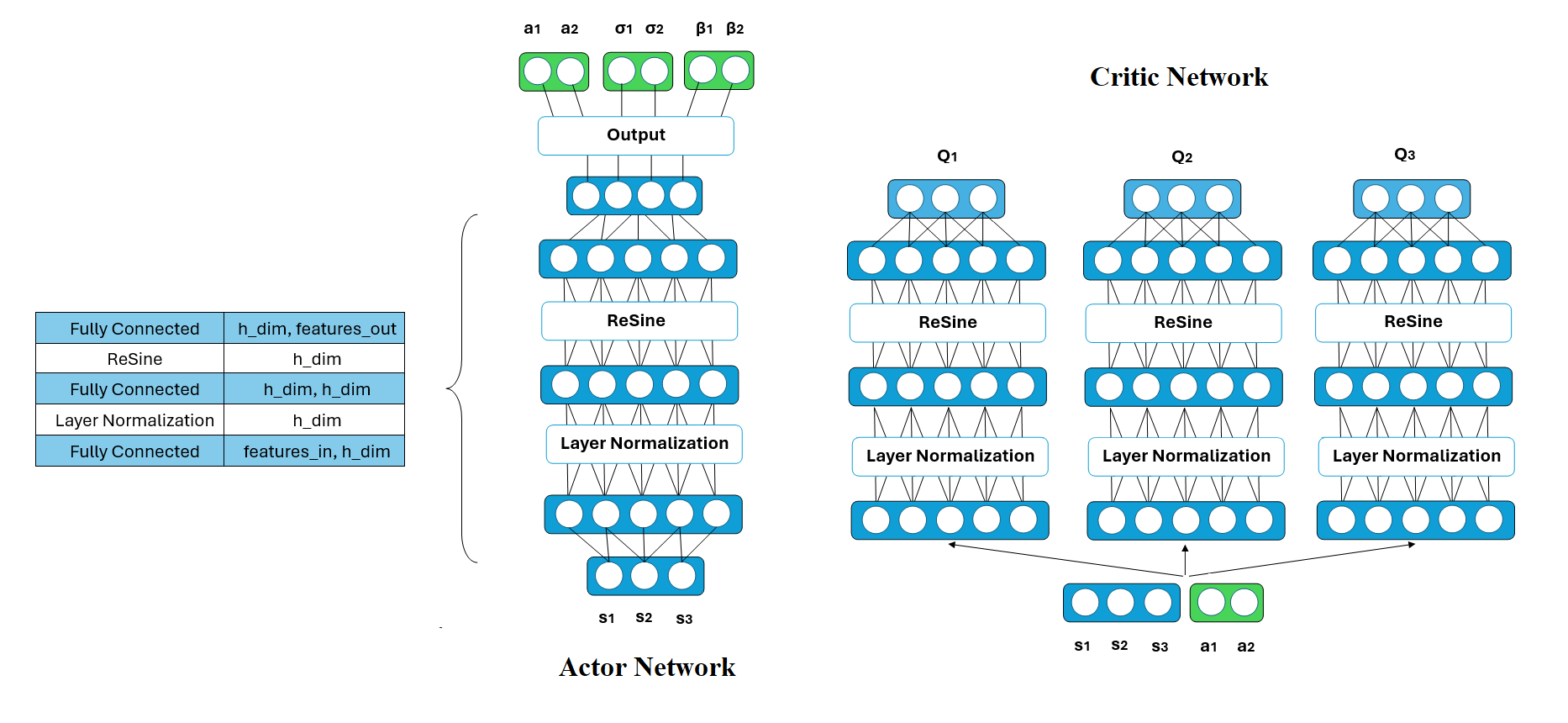}
    \caption[Caption for LOF]{Architecture of the Actor-Critic network}
    \label{fig:Actor_Critic4}
\end{figure*}

Each step during and after Exploration we set new seeds for the libraries we work with (namely, PyTorch, numpy, random).

After the Replay Buffer is filled with the Exploratory data, a new transition enters it at the last index. Then we do roll or shift left operation so that the first (oldest) transition appears at the last index and is ready to be overwritten. When we encounter a transition that contains $\widehat{r_t}=r_{done}$ at the oldest placeholder, we do shift left 2 positions instead of 1. At a step close to the artificial termination limit we zero scaled actions and leave only noise.

An important adjustment to increase speed of training is to place the Replay Buffer on the same device the network's learning is accomplished: we created State, Action, Reward, Done, Next State placeholders with size ($N_{max}$, feature dimension). It is significantly more computationally efficient to put single instance of each feature to the device placeholder than to sample a bigger batch and put it from one device to another at each training step, not to mention that instances can be accessed just by list of indexes. 

\begin{figure}[H]
    \centering
    \includegraphics[width=7cm]{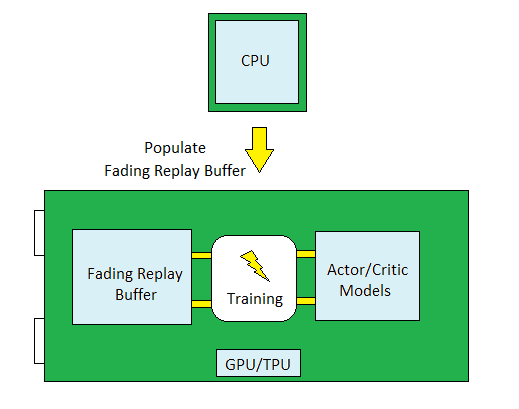}
    \caption[Caption for LOF]{Representation of Device occupation and training}
    \label{fig:filling_frb}
\end{figure}

 Even though we used GPU: NVidia RTX 3000 series, CPU: AMD Ryzen 5..9. for testing, to decrease computational time we recommend performing a PyTorch's trivial just-in-time (JIT) compilation on new modules converting them to C++ optimized graphs including Loss Functions which are processed as modules with forward pass.

All these hyperparameters and measures constitute the default implementation of the algorithm, \textit{Symphony-S3}. Other configurations are possible, which result in different associated parameters. One can unlink the correspondences between parameters completely for related purposes, though we do not recommend a Replay Buffer size lower than 7680*50=384,000.

\subsection*{Sample-efficient version, (SE)}

We made a Sample-efficient configuration by decreasing the noise level $\epsilon \sim a_{max}*1/\pi*\mathcal{N}(0.0,\,1.0)$, where $\mathcal{N} \in [-\pi,\pi]$.

\subsection*{Sample-Proximity and Safety, (S2)}
 It is possible to shift focus from Sample-Efficiency to Sample-Proximity and Safety just by increasing the noise level or sending $\beta$ with a higher order than 2.0 which will eventually punish strong action values harder. However, our Full Swaddling Function was balanced while the parametric noise value of 1/e perfectly suits it and already at the highest level. Layer Normalization and Resine Activation Function provide indirect generalization for the Input and Hidden Layers, respectively, whereas the Output Layer was left untouched. Original Dropout Layer\cite{hinton2012improving} did not suit as it would zero some nodes distorting final distribution. 
 
 Last but not least we added a specific implementation of \textit{Gradient Dropout}\cite{gradientdropout}, which instead of zeroing some nodes as in Dropout Q Functions (DroQ)\cite{hiraoka2021dropout}, does not calculate their gradients during backpropagation. This can be achieved simply by (mask is an original dropout mask, $x^*$ - detached values):

\begin{equation}
mask*x + (1-mask)*x^*
\end{equation}

Gradient Dropout after Fully Connected Output Layer in Actor and Critic with a dropout probability of $50\%$ was the last missing ingredient. We, additionally, decreased the learning speed: $\alpha_{lr}=\frac{1}{2}\mathrm{e}{-4}$ and increased initial generalization: $N_{exp}=20,480$, $N_{rb}=25\,N_{exp}=512,000$.

\subsection*{Embedded devices (ED), Baseline Model}
A configuration for less computationally powerful devices like Jetson Series: $N_{exp}=7680$, $N_{rb}=384,000$, $h_{dim}=384$, each Critic outputs $96$ nodes, $\mathcal{B}=288$. Data is stored in Fading Replay Buffer as float16, converted to float32 during the sample retrieval process.

\section{Experiments}

We set our experiments in Humanoid-v4, OpenAI's Gymnasium (Mujoco) Environment. Earlier versions of the Mujoco Humanoid environment, e.g. Humanoid-v1, had a maximum scale of action within -1 and 1, and due to the complexity of agent training, the maximum scale decreased to [-0.4, 0.4]. We returned the limit of the scale value to 1.0, since we have internal regularization. We did 10 random seed experiments for following models: Symphony-S3, Symphony-SE , Symphony-S2, Symphony-ED (Baseline). Symphony-S2 uses 20,480 exploratory steps, while others use 10,240. Total number of steps $3*10^6$, the episode limit was 1000 steps. We took readings in two ways, step-wise, when we sampled 25 trials each 2500 steps to estimate return and episode-wise to read an average $\sigma$ value in batch (Exploratory steps and episodes were omitted on the graphs and tables).  

As was said for the Baseline model we choose Symphony-ED which to some extent close to Soft-Actor and Critic algorithm in performance but still generates alternating leg movements, though upper-body can stay underdeveloped, variance is relatively low:

\begin{figure}[H]
    \centering
    \includegraphics[width=8cm]{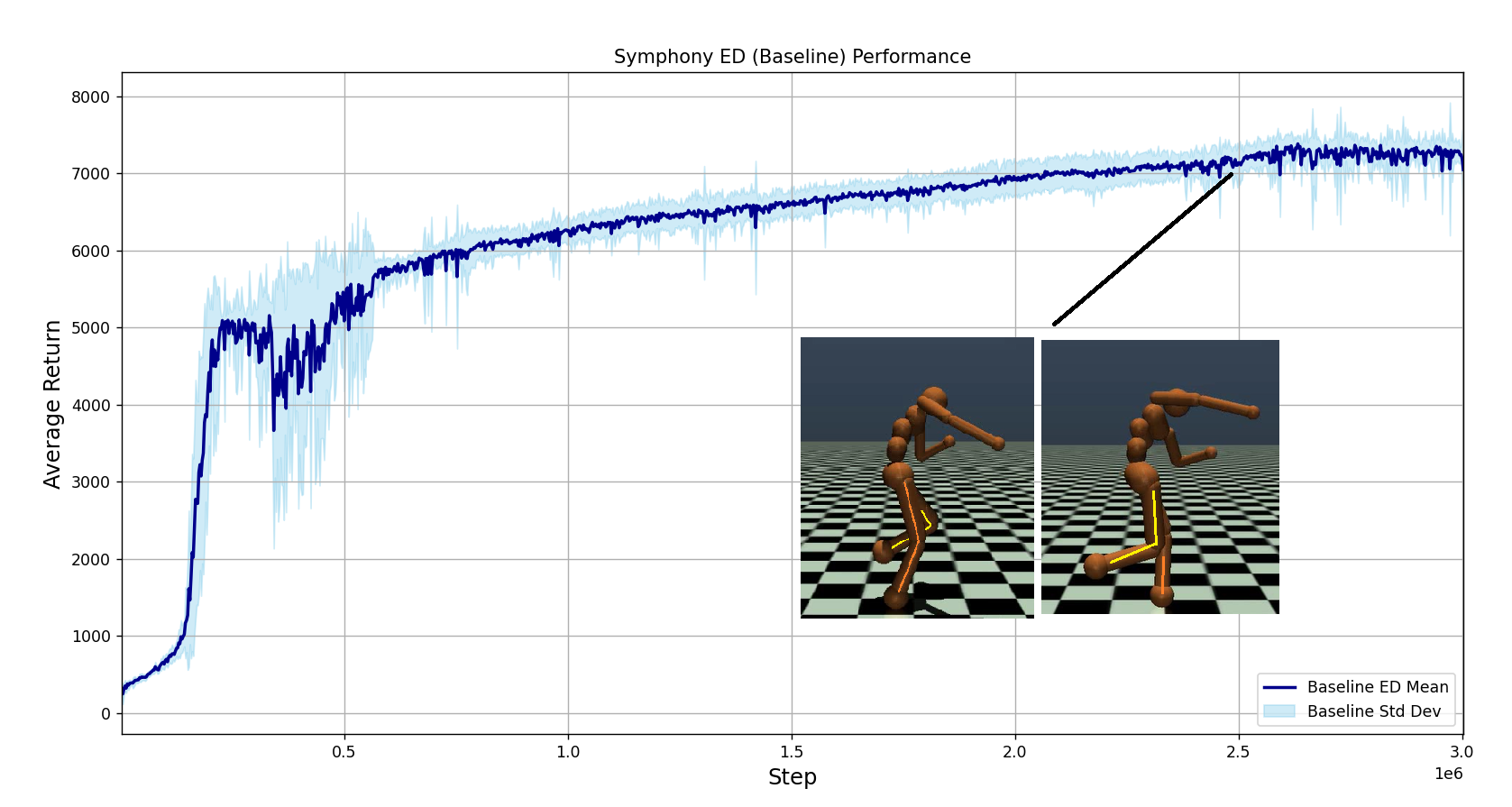}
    \caption[Caption for LOF]{Symphony-ED Return step-wise (orange lines depict supporting leg at a point in time).}
    \label{fig:ED}
\end{figure}

In contrast, Sample-Efficient version, Symphony-SE develops a better gait, but suffer from high volatility especially when the agent's speed increases. It would be wise to decrease learning rate after some point.

\begin{figure}[H]
    \centering
    \includegraphics[width=8cm]{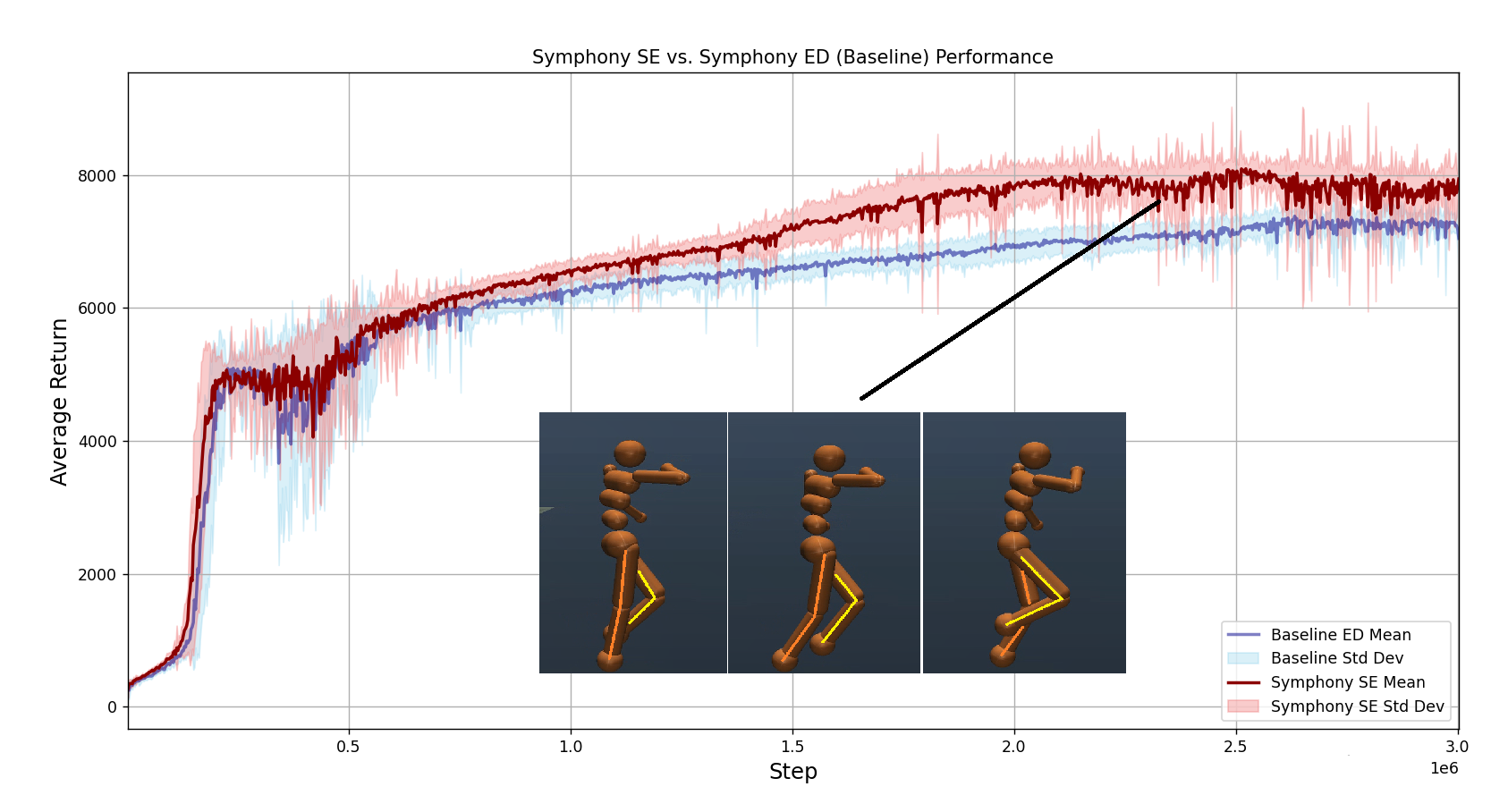}
    \caption[Caption for LOF]{Symphony-SE Return step-wise (orange lines depict supporting leg at a point in time)}.
    \label{fig:SE}
\end{figure}

Symphony-S3 shows better development through the entire training, though as S3 is very close to SE model, it still suffers of higher variance at higher speeds. The same technique of decreasing learning rate can work here as well.

\begin{figure}[H]
    \centering
    \includegraphics[width=8cm]{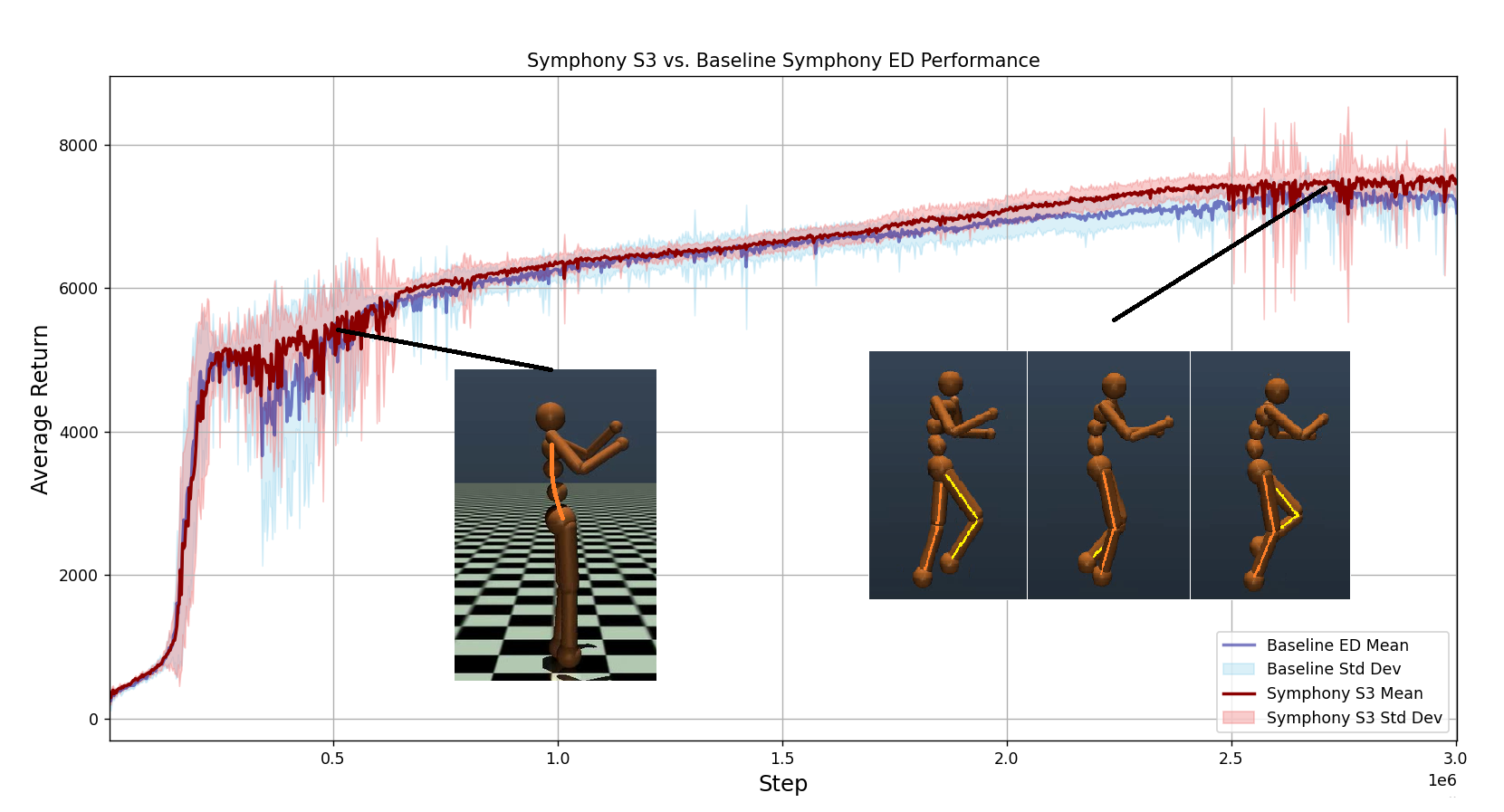}
    \caption[Caption for LOF]{Symphony-S3 Return step-wise (orange lines depict supporting leg at the moment)}.
    \label{fig:S3}
\end{figure}

Symphony-S2 solves most problems of SE and S3 algorithms in exchange of sample-efficiency, however, we can see that the average Return shows positive dynamics, while for the Baseline model it might hit plateau.

\begin{figure}[H]
    \centering
    \includegraphics[width=8cm]{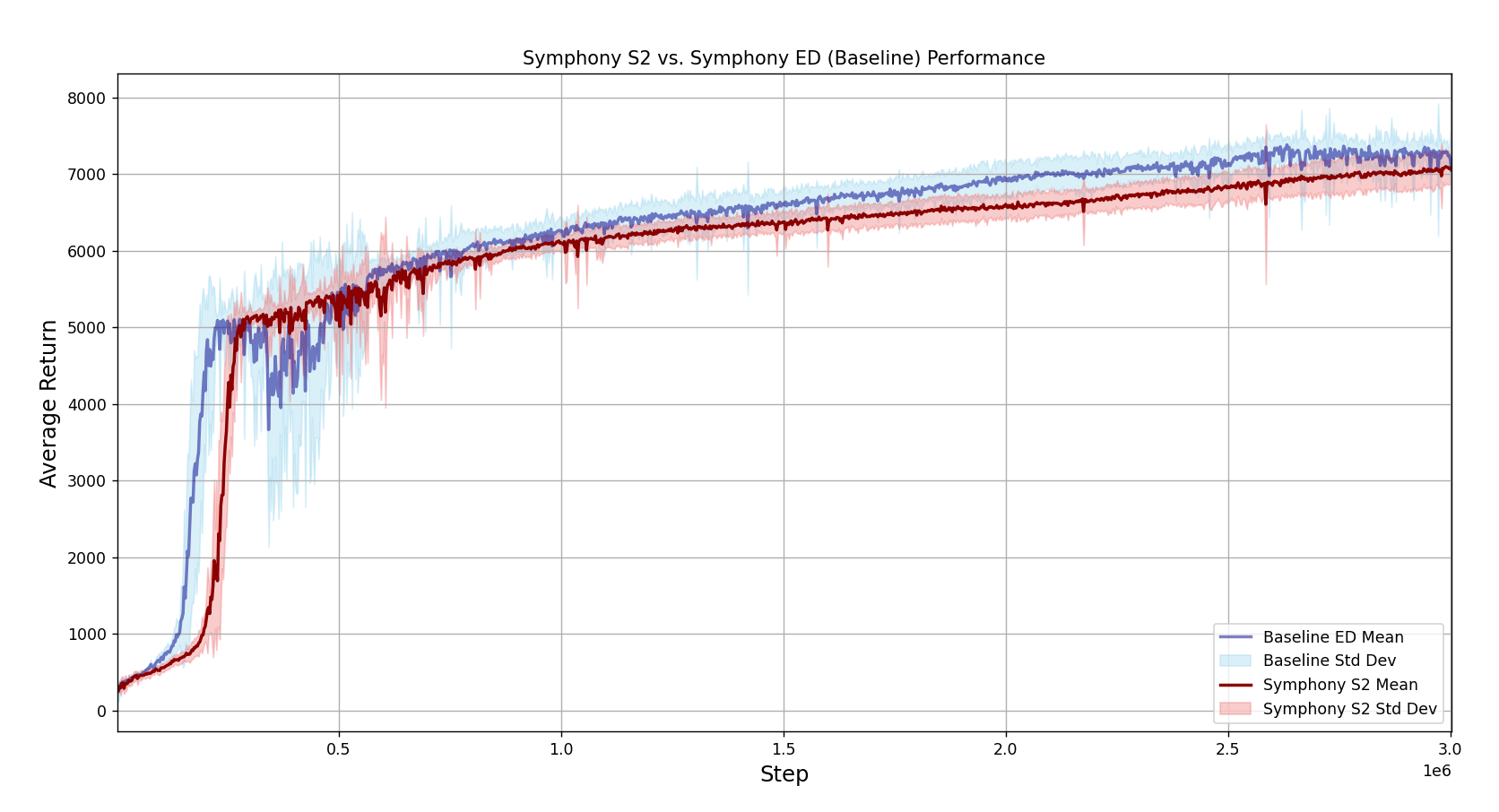}
    \caption[Caption for LOF]{Symphony-S2 Return step-wise}.
    \label{fig:S2}
\end{figure}

The diagram below shows scaling factor behavior episode-wise. For each algorithm, we trimmed the values for the first episode when the number of steps reached $3*10^6$. It can be seen that the Scaling factor initially varies greatly (least of all for Symphony-S2), but then decreases closer to $\frac{1}{e}$, and slowly increases from this value.

\begin{figure}[H]
    \centering
    \includegraphics[width=8cm]{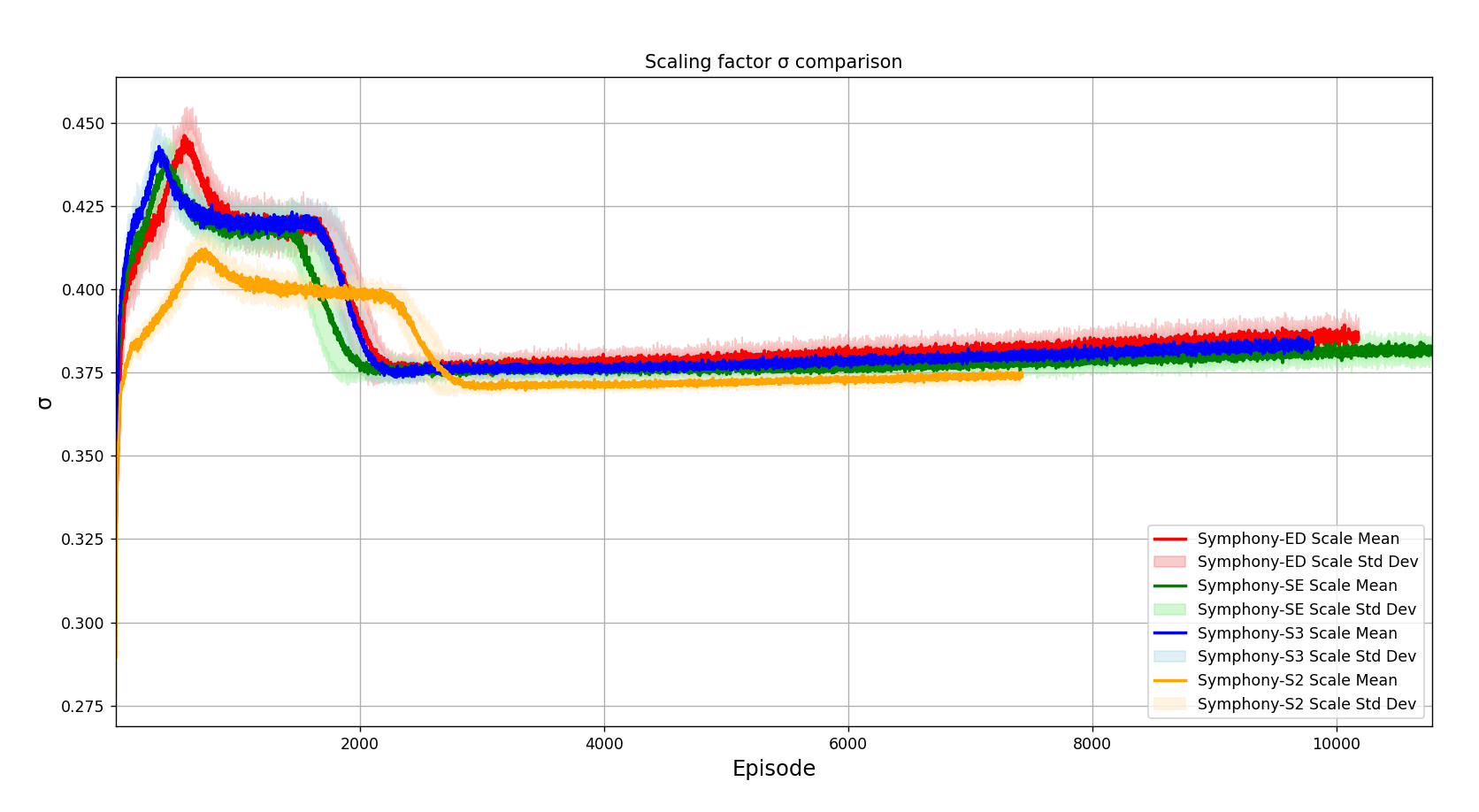}
    \caption[Caption for LOF]{Scaling Factor comparison.}.
    \label{fig:Scaling_factor_comparison}
\end{figure}

We were interested in finding out the average value for the episode when $3*10^6$ steps are reached. A preliminary conclusion can be drawn that, better generalization and slower learning rate result in better prediction as the agent falls less:

\begin{table}[htbp]
\caption{Episode number when the $3*10^6$ steps were reached}
\begin{center}
\renewcommand{\arraystretch}{1.2}
\begin{tabular}{ m{3em}|m{10em}  }

  Version & Average episode\\ 
 \hline
 ED & $11530.5\pm1132.8$  \\
 SE & $12433.5\pm1055.0$  \\
 S3 & $11411.5\pm966.0$  \\
 S2 & $\mathbf{8284.1\pm753.7}$  \\

\multicolumn{2}{c}{Symphony-S2's accurate gait during training}\\
\multicolumn{2}{c}{\includegraphics[width=0.7\linewidth]{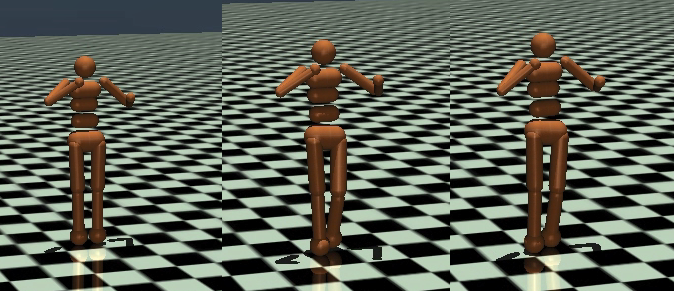}}\\
\hline
 
\end{tabular}
\end{center}
\end{table}

However, this does not imply that top scores are reached faster. To develop human-like movements one may need to sacrifice Sample-Efficiency to some extent.

\begin{figure}[H]
    \centering
    \includegraphics[width=6.2cm]{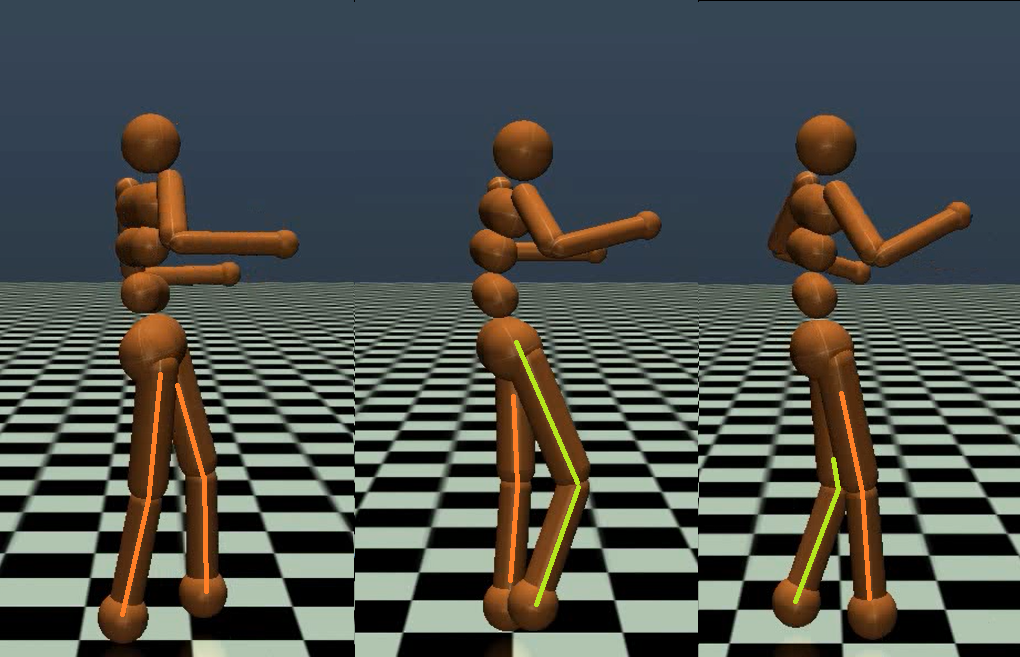}
    \caption[Caption for LOF]{Depiction when agent uses both hands and legs for balance during training}
    \label{fig:Humanoid}
\end{figure}

\begin{table}[htbp]
\caption{Step when the specific threshold of average Return is reached}
\begin{center}
\renewcommand{\arraystretch}{1.2}
\begin{tabular}{ m{5em}|m{3em}|m{3em}|m{3em}|m{3em}  }

  & \multicolumn{4}{|c}{The Symphony version } \\
 \hline
  Avg. Return & ED & SE & S3 & S2\\ 
 \hline
 5000 & 227k & \textbf{217k} & 240k & \textit{285k} \\

 6000 & 782k & \textbf{657k} & 682k & \textit{880k}\\

 7000 & 2,060k & \textbf{1,392k} & 1,895 & \textit{2,768k}\\

\end{tabular}
\end{center}
\end{table}

Even with all this, there's no $100\%$ guarantee that there will be absence of developmental defects. In our world, external obstacles often play an additional regulatory role. Following the logic of gradual human learning, we recommend starting training without obstacles, then gradually increasing the difficulty, filling a new Fading Replay Buffer with new exploratory data. 

\subsection*{\textbf{Symphony-$\Tilde{S2}$}}

For practical use and for the upcoming ablation study, we decided to stick with Symphony-S2, with some modifications. For continuous learning with terminal cases to prevent overflow, we subtract a small value of $\epsilon$ from the "done=1.0" feature, and then shifting it by 2 positions. If this value is below some critical threshold close to 1.0, we perform a standard shift by 1 position. This will ensure that after the N-th cycle, the terminal transition is overwritten. We additionally optimized the Fading Replay Buffer by refactoring the resource-consuming roll operation into a circular buffer with a cyclic pointer, while also implementing in-place operations to reduce memory overhead.

Instead of the modified hyperbolic tangent for the fixed weights, we used a bell-shaped Gaussian with a power of $2\pi$, a width of $1/\phi$, and a shift of $0.5 + 0.5 \, (1/\phi) = \phi/2$, which has a maximum at the shift point $\phi/2$.

\begin{table}[!ht]
\caption{Fixed weights}
\centering 
\begin{tabular}{c|c}

\includegraphics[width=0.36\linewidth]{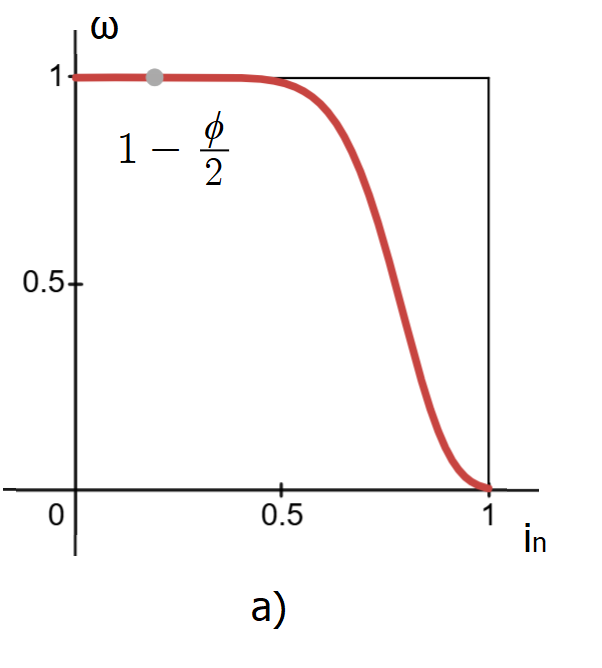} & \includegraphics[width=0.36\linewidth]{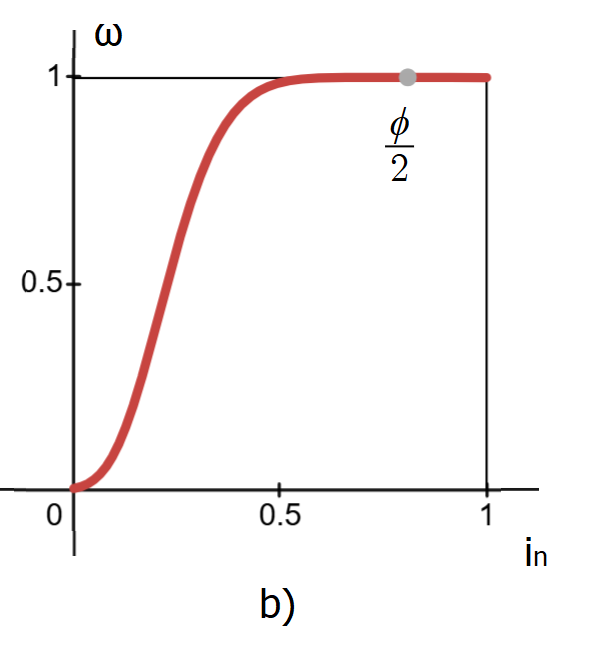} \\

a) for Target Critic &  b) for Fading Replay Buffer \\
\\
$\omega_{i_n} = e^{-(\frac{|1-b-i_n|}{s})^{p}}$
 &
 $\omega_{i_n}=e^{-(\frac{|i_n-b|}{s})^{p}}$\\
\\
\hline
\\
\multicolumn{2}{c}{$b=\phi/2$, $s=1/\phi$, $p=2\pi$}

\end{tabular}
\end{table}

When using new weights for Target Critic, we reduce the importance of minimum values, while when using them for Fading Replay Buffer, our algorithm will not jump towards the latest transitions immediately.

Instead of 50\% dropout probability or p=0.5, we used random probabilities (random values were generated using Gaussian Normal Distribution processed through Sigmoid function).

\begin{table}[ht]
\centering
\begin{tabular}{c}

\\
\normalsize
1) $ x \leftarrow X\sim\mathcal{U}(0,1)$ \\[1.5em]
\normalsize
2) $p \leftarrow sigmoid(P\sim\mathcal{N}(0,1))$ \\[1.5em]
\normalsize
3) $mask \leftarrow \mathbb{I}(x > p)$ \\[1.5em]
\end{tabular}
\end{table}

parameters: Normal Gaussian Noise $N=1/e^2*\mathcal{N}(0.0,\,1.0)$, $\mathcal{N} \in [-e, e]$, $\mathcal{G}=3$, $\alpha_{lr}=1\mathrm{e}{-5}$, $N_{exp}=20,480$, $N_{rb}=50$, $N_{exp}=1,024,000$,  $h_{dim}=N_{rb}\div2000=512$, $\mathcal{B}=384$.

\section*{Conclusion}

In our work, we traded off a fast convergence rate for the sake of eliminating jerky motions and convergence to suboptimal policies during training. We acknowledge that our results might not achieve state-of-the-art benchmarks in any single direction as the Symphony algorithm is the result of rigorous work balancing three constraints - Sample Efficiency, Sample Proximity and Safety of Actions (regarding extreme action values) for practical applications. To describe the Symphony Algorithm in a nutshell, it can be considered as a biological organism where all parts complement each other: disabling one weakens another. Even though the update-to-data ratio is 3 it is mostly used for coherent understanding of the distribution shift. We conducted hundreds of experiments before finding suitable fixed parameters for humanoid robots. In the next article, we'd like to conduct a full analysis of each component separately, we are preparing an ablation study as a separate project.

\section*{ACKNOWLEDGMENTS}

I want to thank my Lord, Jesus Christ. In the scientific community, this name, and religion in general, is not particularly welcome. I consider it important to say that I cannot even breathe without Him. He gives us life and the opportunity to create and build, He changes our character for the better. He gave me many ideas not when I spent hours on end sitting at the computer, but when I finally decided to spend time with my family, or when I listened to a professor. He teaches me to value, respect, and love my closest ones.

This research was funded by the European Union project RRF-2.3.1-21-2022- 00004 within the framework of the Artificial Intelligence National Laboratory and project TKP2021-NVA-09, implemented with the support provided by the Ministry of Innovation and Technology of Hungary from the National Research, Development and Innovation Fund, financed under the TKP2021-NVA funding scheme.

We gratefully acknowledge the support of the project “Designing an Energy-efficient Full-sized Humanoid Robot with Fast Adaptive Model-based Neuromorphic Control Architecture”, grant number $\#201223FD8812$.

We are grateful for provided resources and facilities to Nazarbayev University and University of Szeged.

We also acknowledge the help of AI tools such as Gemini/ChatGPT/DeepSeek for implementations of different functions.

\bibliographystyle{IEEEtran}
\bibliography{bibliography}

\end{document}